\pdfoutput=1

\documentclass[11pt]{article}

\usepackage[]{acl}

\usepackage{times}
\usepackage{latexsym}
\usepackage{float}
\usepackage[normalem]{ulem}
\useunder{\uline}{\ul}{}
\usepackage[T1]{fontenc}

\usepackage[utf8]{inputenc}

\usepackage{microtype}

\usepackage{multirow}
\usepackage{amsmath}
\usepackage{graphicx}
\usepackage{caption}
\usepackage{subcaption}
\usepackage{xcolor}
\usepackage{hyperref}
\definecolor{targetcolor}{HTML}{00CC00}
\definecolor{datesamplecolor}{HTML}{00994D}
\definecolor{citysamplecolor}{HTML}{7EA6E0}
\definecolor{intentsamplecolor}{HTML}{A680B8}
\newcommand{\deemph}[1]{{\color{black!40}#1}}

\usepackage{booktabs}

\title{Grounding Description-Driven Dialogue State Trackers with Knowledge-Seeking Turns}

\author{Alexandru Coca$^{\dagger}$, Bo-Hsiang Tseng$^{\ddagger}$, Jinghong Chen$^{\dagger}$, Weizhe Lin$^{\dagger}$,\\ \textbf{Weixuan Zhang}$^{\dagger}$, \textbf{Tisha Anders}$^{\dagger}$$^*$,  \textbf{Bill Byrne}$^{\dagger}$ \\
        ${}^\dagger$Department of Engineering, University of Cambridge, United Kingdom \\
        ${}^\ddagger$Apple \\
        $^{\dagger}$\texttt{\{ac2123, jc2124, wl356, 
wz315, wjb31\}@cam.ac.uk} \\
        $^{\ddagger}$\texttt{bohsiang\_tseng@apple.com}~~~~~~ $^*$\texttt{anderstisha@gmail.com} }

\begin{document}
\maketitle
\begin{abstract}
Schema-guided dialogue state trackers can generalise to new domains without further training, yet they are sensitive to the writing style of the schemata. Augmenting the training set with human or synthetic schema paraphrases improves the model robustness to these variations but can be either costly or difficult to control.  We propose to circumvent these issues by grounding the state tracking model in knowledge-seeking turns collected from the dialogue corpus as well as the schema. Including these turns in prompts during finetuning and inference leads to marked improvements in model robustness, as demonstrated by large average joint goal accuracy and schema sensitivity improvements on SGD and SGD-X\footnote{Our code will be released upon publication.}.
\end{abstract}

\section{Introduction}
Task-oriented dialogue (TOD) agents provide natural language interfaces that users can interact with to access a wide variety of services, from airline search \citep{seneff2000dialogue} to complex customer service applications \cite{DBLP:conf/naacl/ChenCYLY21}. To enable this, agents track key information communicated by the user as the conversation progresses. This is known as \textit{dialogue state tracking} (DST). Commonly, the dialogue state is represented as a sequence of task-specific \textit{slot-value pairs}\footnote{For example, for a restaurant booking a sequence could be \textit{day=friday, time=7pm, guests=1, restaurant=nandos.}}.

A common DST assumption is that the set of slots and values a user may communicate, the \textit{domain ontology}, is known at design time. Hence, extensive data collection and annotation is needed to support new domains, which hinders the scalability of this approach. \citet{DBLP:conf/aaai/RastogiZSGK20} address this issue by creating the schema-guided dialogue dataset (SGD). In SGD, the information available to a TOD agent is organised as \textit{schemas}\footnote{See schema examples here: \href{https://bit.ly/3RJ6u4l}{https://bit.ly/3RJ6u4l}.} describing \textit{services} with which users can interact. Each service has \textit{user intents} representing the tasks users can complete by interacting with the agent (e.g. \textit{find restaurants}). Several slots are associated with each intent and the schema provides a \textit{natural language description} for each intent and slot. The insight motivating  {\em description-driven DST} is that these descriptions alone can be used in tracking the dialogue state in a form close to natural language. This has emerged as a powerful approach for few-shot and zero-shot DST \cite{jacqmin-etal-2022-follow} and benefits from recent advances in language modelling. For example, \citet{DBLP:journals/corr/abs-2201-08904}  finetune T5 \citep{DBLP:journals/jmlr/RaffelSRLNMZLL20} to generate the dialogue state conditioned on the dialogue history and a descriptive prompt containing all intent and slot descriptions in a service schema. The use of natural language in the descriptive prompt enables the underlying language model to generalise to new services, whose schemas are not seen in training.

While the reliance on natural language is a strength, \citet{DBLP:conf/aaai/0001GR0ZW22} show that state-of-the-art (SOTA) schema-guided DST models are not robust to the style of descriptive prompts: in SGD, the training schema contains a single description per slot or intent, and models trained with prompts composed from this schema alone are prone to overfitting. \citet{DBLP:conf/aaai/0001GR0ZW22} show this limitation can be mitigated by increasing prompt diversity. They use a large number of human annotators alongside expert curation to create  diverse schema paraphrases that are used for model robustness improvement. This is a costly process that is not easily scalable. 
\begin{figure*}[]
    \centering
    \includegraphics[width=0.85\textwidth]{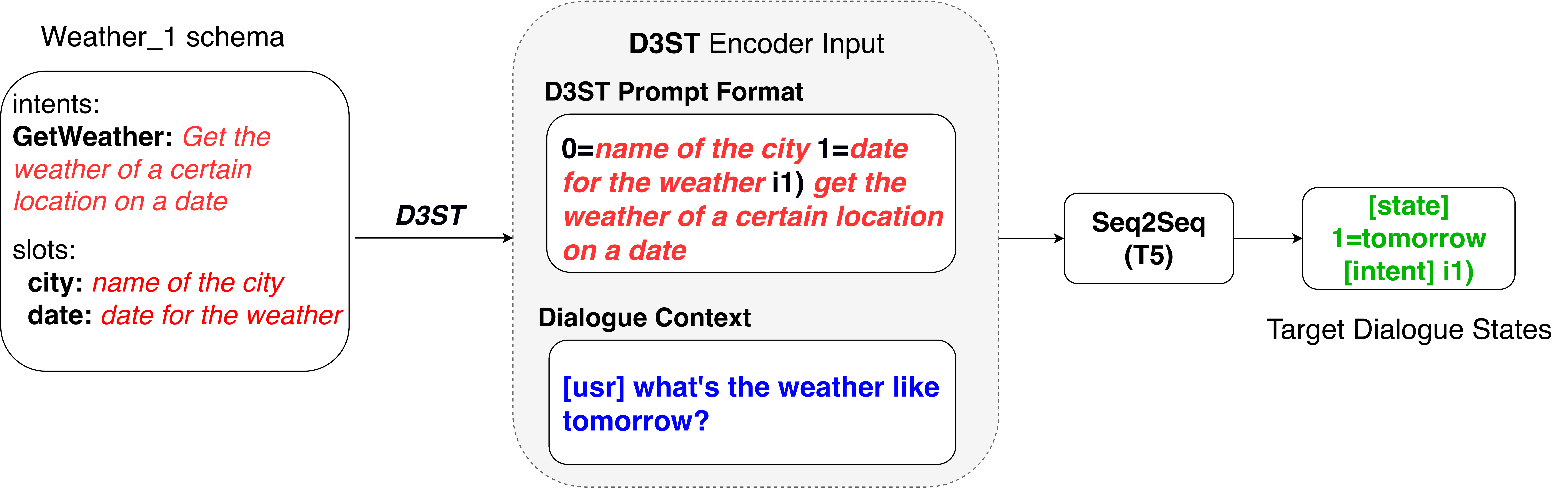}
    \caption{D3ST input and target format. On the left, we show a schema excerpt, where \textbf{slot} and \textbf{intent} names, in bold face, are followed by their \textit{natural language description}. The encoder input, represented in the centre, is a string, the concatenation of two elements: the \textit{prompt} which describes what information should be output and  the \textit{dialogue context} from which the information should be extracted. On the right we show the target dialogue state.}
    \label{fig:D3STFigure}
\end{figure*}

As an alternative to additional human annotation of the schema, we show that the SGD training dialogues themselves exhibit sufficient diversity of expression such that they can be used to overcome the lack of diversity in the SGD schema descriptions. We {\em ground} DST prompts in the dialogue context by concatenating the schema descriptions with dialogue turns extracted from the SGD corpus based on similarity to the dialogue state. We find that this approach is more effective than using synthesised prompts and even outperforms or is comparable to the highly-curated human-written prompts used by \citet{DBLP:conf/aaai/0001GR0ZW22}, when evaluated with medium and large language models. We evaluate our methods using the SOTA D3ST prompting scheme \cite{DBLP:journals/corr/abs-2201-08904} on the SGD and SGD-X \cite{DBLP:conf/aaai/0001GR0ZW22} datasets.

\section{Related work}

Neural classification is effective for DST  when the domain ontology is fixed and known at design time \cite{DBLP:conf/acl/MrksicSWTY17}. Adapting such models to track new slots and domains requires annotated conversational data and thus data scarcity is a long-standing issue in DST research \cite{jacqmin-etal-2022-follow}. Scarcity has been addressed by copy-enhanced generation \citep{wu2019transferable}, reading comprehension \citep{gao2019dialog} and adapting pretrained language models to generate the state given the dialogue context alone \citep{peng2020soloist, hosseini2020simple}. These were improved upon by transfer learning from question-answering tasks \cite{DBLP:conf/emnlp/LinLMMZCWYCSF21}, which in turn was outperformed by schema-guided models \citep{DBLP:conf/emnlp/00010O21, DBLP:journals/corr/abs-2201-08904, lin-etal-2021-leveraging}. Recently, \citet{DBLP:conf/naacl/Gupta0Z0RW22} apply in-context tuning \cite{DBLP:conf/emnlp/MinLHALHZ22} to DST, creating training prompts which contain a dialogue and its target state sequence. Their model thus learns from DST task demonstrations.

\citet{DBLP:conf/aaai/0001GR0ZW22} investigate the robustness of SOTA schema-guided dialogue state trackers to schema changes. This is a new line of research, as previous work concerns other robustness issues that generally affect DST, such as variations in the conversational data distribution, noise, and adversarial perturbations \cite{jacqmin-etal-2022-follow}. Through extensive, crowdsourced\footnote{The SGD schema were rewritten by over 400 annotators and curated by dialogue experts, over the course of one month.}, schema paraphrase collection, \citet{DBLP:conf/aaai/0001GR0ZW22} report that DST performance degrades substantially when  models trained on one set of prompts are evaluated on manually paraphrased prompts. By contrast, \citet{DBLP:conf/naacl/CaoZ21} report little degradation in DST with backtranslated prompts, suggesting that backtranslation is a weak proxy for actual human variability. \citet{DBLP:conf/aaai/0001GR0ZW22} perform data augmentation (DA) for robust DST, finding that prompts obtained via automatic paraphrasing lag in quality relative to manual paraphrases. Ours is the first work to address the gap between synthetic methods, such as backtranslation, and manual paraphrasing. We show that the gains from manual paraphrasing can be achieved by mining the existing annotated dialogues used for training the DST model in the first place. 

\section{Robust DST with grounded prompts}

We review D3ST, a SOTA description-driven DST model (Section \ref{sec:d3st-tutorial}). We then describe our grounding method that extracts turns from the corpus (Section \ref{sec:turn-mining}) and uses them to design prompts for robust DST with D3ST (Sections \ref{sec:grounded-training} \& \ref{sec:gpe}).

\subsection{Description-driven dialogue state tracking}
\label{sec:d3st-tutorial}

Figure \ref{fig:D3STFigure} shows the inputs and outputs of D3ST \cite{DBLP:journals/corr/abs-2201-08904}. The model is implemented with T5, an encoder-decoder language model \cite{DBLP:journals/jmlr/RaffelSRLNMZLL20}. The encoder input, represented in the centre, comprises a \textit{prompt} describing what information should be tracked by the DST model, and the \textit{dialogue context}, a conversation between a user and an agent from which slot-value pairs should be  extracted. The prompt is a concatenation of slot and intent descriptions, extracted from the service schema (on the left). Each description is prefixed by a randomly assigned index prior to concatenation. \citet{DBLP:journals/corr/abs-2201-08904} motivate their use of random indices to replace slot and intent names because names convey little semantic information and may be ambiguous\footnote{For example, the \textit{location} slot name may be used to refer to both a city name and an address across different services.}. In this paper, we will refer to this \textit{prompt format} as \textbf{D3ST}.

The model is trained to output index-value pairs for all slots mentioned in the conversation (i.e. the \textit{active slots}) as well as an index representing the active intent. These are represented on the right in Figure \ref{fig:D3STFigure}. The slot-value pairs mentioned in the conversation can be recovered by replacing the predicted indices with their corresponding slot names. The user active intent is found by replacing the predicted index with the name of the intent.

\subsection{Mining turns for prompt design}
\label{sec:turn-mining}

\begin{table}[htpb]
\centering
\small
\begin{tabular}{ll}
\toprule
1. \texttt{REQUEST(restaurant\_name)}   \\ 
SYS: Where do you want to dine?           \\
\hline
2. \texttt{INFORM(restaurant\_name=Nandos)} \\
USR: I want Nando's. \\
\hline
3. \texttt{INFORM\_INTENT(find event)} \\ USR: What shows are on?                   \\
\hline
4. \texttt{OFFER\_INTENT(buy ticket)}  \\ SYS: Want tickets?                        \\
\bottomrule
\end{tabular}
\caption{Sample semantic annotations}
\label{tab:semantic}
\end{table}
\begin{figure*}[htpb]
    \centering
    \includegraphics[width=\textwidth]{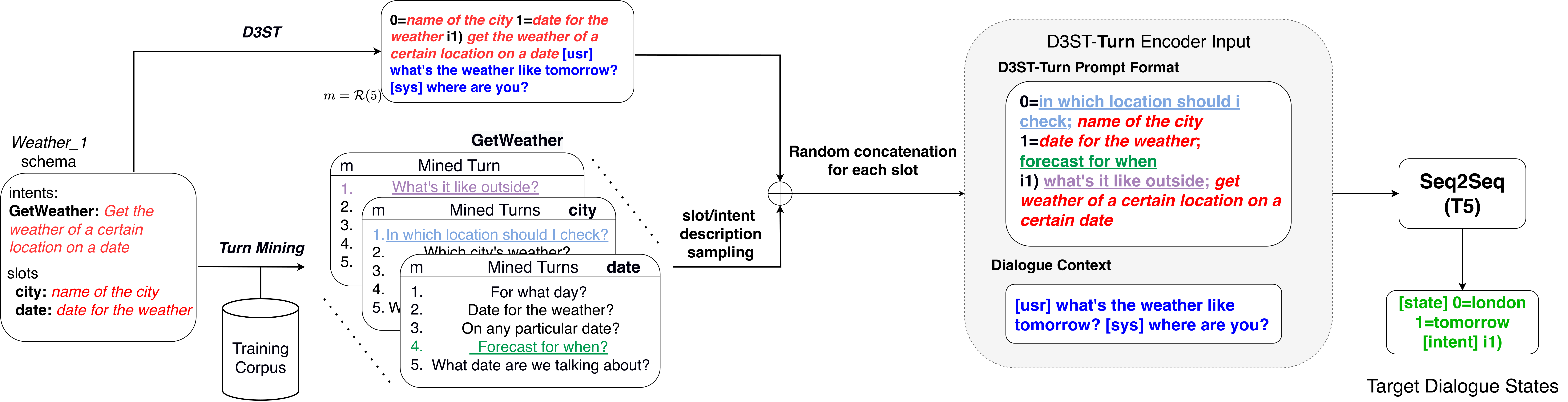}
    \caption{Visual representation of D3ST-Turn prompting. Underlined knowledge seeking turns are those chosen at random for inclusion in the sample D3ST-Turn prompt shown.}
    \label{fig:Turn}
\end{figure*}
We now discuss how to extract turns from the corpus to design better prompts. Our approach involves an automatic step that uses the semantic annotations in the corpus followed by a verification step to ensure that the turns selected are diverse.

Each turn in SGD is semantically annotated with one or more \textit{dialogue actions} which describe what is being communicated (Table \ref{tab:semantic}). We focus on \textit{knowledge-seeking} turns (KSTs). These are annotated with a single \texttt{REQUEST} \textit{dialogue act} and associated with a \textit{single} slot, without a value mention (Table \ref{tab:semantic}, line 1). Selecting turns annotated with a single slot allows us to unambiguously associate  them with slots in the D3ST prompt. We do not mine \textit{informational} turns (labelled with \texttt{INFORM}) since these mention a specific, known value, often without reference to the underlying slot (Table \ref{tab:semantic}, line 2). Such turns could be combined with schema information to form \textit{exemplar-based} prompts as done by (Figure 1 in \citet{DBLP:conf/naacl/Gupta0Z0RW22}), a more complex approach which we discuss in Section \ref{sec:comparisons}.

To select turns for a given slot, $s$, we filter the corpus to get all the knowledge-seeking turns relating to it. We manually select $5$ of these, repeating this process for each slot in every service in the training data. See examples in Table \ref{tab:kst-example}. In a similar fashion, we select $5$ turns from those labelled with a single \texttt{INFORM\_INTENT} or \texttt{OFFER\_INTENT} act (Table \ref{tab:semantic}) for every intent in the training schema. 
\begin{table}[htbp]
\centering
\small
\begin{tabular}{cc}
\toprule
Index & Selected Knowledge-seeking Turn                           \\ \hline
1      & Which event are you looking to book     \\
2      & Do you have any particular show in mind \\
3      & And what is the event                   \\
4      & What event do you wish to see           \\
5      & What is the event you are looking for   \\ \bottomrule
\end{tabular}
\caption{Selected turns of the \textit{event name} slot}
\label{tab:kst-example}
\vspace{-0.2cm}
\end{table}

We opt to select the turns manually because our goal is prompt diversity. Our SGD analysis revealed that the knowledge-seeking turns tend to  be biased towards specific vocabulary and syntactic patterns. For example, among the $173$ turns in which the user requests the price of a rental car, $71.1\%$ contain the word \textit{cost}, $42.8\%$ contain the word \textit{total} and, $27.7\%$ contain \textit{total cost}. In contrast, \textit{price} appears in just $11.0\%$ of the turns.

All turns were mined by one student in one day, despite SGD being the largest schema-guided TOD corpus. In practice, schemas are induced by developers from unlabelled conversation databases \cite{DBLP:conf/naacl/YuW0SSS22}. The turns could be collected as part of this process with negligible overhead. We handle slots with few KSTs as described in Appendix \ref{appdx:kst-details}.

\subsection{Grounding prompts}
\label{sec:grounded-training}
To ground a schema description in its conversational use, we concatenate it with randomly selected knowledge-seeking turns from the mined collection (Figure \ref{fig:Turn}). In the example shown, the sampled knowledge-seeking turns for the \textit{city} and \textit{date} slots are \textit{In which location should I check?} and \textit{Forecast for when?}, respectively. These are concatenated with the original SGD schema descriptions \textit{name of the city} and \textit{date for the weather} to ground the prompt. We concatenate the turns and descriptions in random order, to prevent the model learning to attend preferentially to one source of information over another. We refer to D3ST trained with prompts grounded in knowledge-seeking turns as {\bf D3ST-Turn} in what follows.

Slot names may provide additional information about the meaning of a slot, so we propose to ground the prompt both in knowledge-seeking turns and slot names. We refer to D3ST trained with prompts grounded in knowledge-seeking turns and slot names as {\bf D3ST-TurnSlot}.

\subsection{Grounded prompt ensembling}
\label{sec:gpe}
Multiple knowledge-seeking turns are available for decoding, enabling us to create multiple \textit{prompt variants}. A given model generates the dialogue state when conditioned on each of these prompt variants, in turn. The state hypothesis is the most commonly predicted string when our \textit{single} model is prompted with the prompt variants. We call this technique \textit{grounded prompt ensembling} (GPE).

\section{Experiments}
\label{sec:experiments}
\subsection{Datasets and metrics}
\label{sec:dataset-metrics}
\textbf{SGD} \cite{DBLP:conf/aaai/RastogiZSGK20} The training set contains $21, 106$ dialogues across $16$ domains. The test set contains $4,201$ conversations, $77\%$ of which have a turn span where the user talks to the agent to access a service unseen in training. $6$ schemas are seen in training whereas $15$ are \textit{unseen}. Hence, this benchmark primarily tests the ability of DST models to accommodate values, slots, prompts and domains it has not been trained on.

\textbf{SGD-X}  \citet{DBLP:conf/aaai/0001GR0ZW22} created SGD-X because they found the linguistic patterns of the SGD unseen services schemata to be too similar to those of the seen schemata\footnote{For example, descriptions of slots with "true" or "false" values always start with \textit{Boolean flag indicating}.}. They use crowdsourcing and dialogue experts to create five \textit{schema variants}\footnote{See examples here: \href{https://bit.ly/3Ev0KrV}{https://bit.ly/3Ev0KrV}.} which are increasingly stylistically and lexically divergent from the SGD schema. A schema variant describes the same services as the SGD schema but with increased linguistic variation. The variants are ordered by the Jaccard distance between the descriptions of the original SGD schemas and the schema variant descriptions. The \textit{v1} variant is the closest while \textit{v5} is the most dissimilar to SGD. Ideally, a robust model should output the correct state regardless of which schema variant is used for prompting. 

\textbf{Metrics} Joint goal accuracy (JGA)\footnote{We use the offical evaluator: \href{https://bit.ly/3B7jD1c}{https://bit.ly/3B7jD1c}} is the percentage of turns where all the slot-value pairs from a given service are correctly predicted. On SGD, it is computed over seen and unseen services. The presence of the unseen services measures the ability of the DST model to make correct predictions for unseen slots and values and to interpret descriptions unseen at training time \cite{DBLP:conf/aaai/RastogiZSGK20}.

For SGD-X, we report the JGA broken down by seen and unseen services and their combination. The JGA coefficient of variation across the five schema variants is termed \textit{sensitivity} ($\mbox{SS}$). It measures how well the model accommodates linguistic variation. Evaluation on the seen portion involves prompting with \textit{paraphrases} of schemata seen in training. Performance decreases if the model overfits to the training descriptions. In evaluation on the unseen portion, the model is prompted with five distinct human-written prompts of increasing dissimilarity to the original SGD. This evaluates if generalisation is robust to linguistic variation.

\subsection{DST models}
\label{sec:motivate-exp}

Our baselines are three D3ST models\footnote{We use T5-base (220M) for all models except in Sec. \ref{sec:scalability}} trained with large augmented datasets. For every training example that uses the D3ST prompt format (Figure \ref{fig:D3STFigure}) linearised from the SGD schema, $k$ additional training examples are created either using synthetic prompts or the $k=5$ SGD-X schemata. We create augmented datasets using three methods, explained below. See Appendix \ref{appendix:state-tracking} for implementation details.

\textbf{1. Backtranslation} We follow \citet{DBLP:conf/aaai/0001GR0ZW22} to create $k=3$ schema variants by backtranslating the SGD schema via Chinese, Japanese and Korean with Google Translate. The augmented dataset is 4 times larger than SGD ( $703, 120$ examples). 

\textbf{2. Easy Data Augmentation (EDA)} \cite{DBLP:conf/emnlp/WeiZ19} We create $k=5$ schema variants by applying word-level perturbation to the SGD schema (EDA). Synonym replacement is applied with probability $0.25$ whereas random insertion, deletion and substitution are applied with probability $0.05$. There are $1,054, 680$ training examples. 

\textbf{3. SGD-X} We create $1, 054, 680$ training examples using the $k=5$ human-written SGD-X schemata. Unlike the other baselines, SGD-X-trained models see the human-written paraphrases of the seen test services during finetuning. In all other experiments, \textit{none} of the SGD-X test prompts are seen during training, and so we refer to this experiment as an \textit{oracle}, following \citet{DBLP:conf/aaai/0001GR0ZW22}.

{\bf Grounded D3ST} Instead of augmentation, we propose to ground D3ST in knowledge-seeking turns by finetuning T5 with the Turn (D3ST-Turn) and TurnSlot (D3ST-TurnSlot) prompts (Section \ref{sec:grounded-training}) on a dataset containing $175, 780$ examples (SGD size). At decoding, the same turns grounding the training prompts are used for seen services. For unseen services, we select five turns per slot as described in Section \ref{sec:turn-mining}. For each test example, we construct a prompt with the same format as in training using knowledge seeking turns selected at random, per slot, from the mined collection. This tests model's ability to interpret additional task-relevant information.

\section{Results and discussion}
\label{sec:results}
\subsection{Robustness via data augmentation}

\begin{table}[tb]
\centering
\resizebox{\columnwidth}{!}{
\begin{tabular}{ccccccc}
\toprule
\# &
Model &
  SGD  &
  SGD-X &
  Seen &
  Unseen &
  $\mbox{SS}\downarrow $ \\ \hline
1 & {D3ST}                                    & 69.8                  & 56.5          & 73.6          & 50.8          & 70.1          \\ 
2 & {D3ST + Backtrans. DA} &  {72.1} & 62.2  & {\color[HTML]{000000} {84.0}} & {\color[HTML]{000000} 54.9} &{\color[HTML]{000000} {53.1}}   \\
 3 & D3ST + EDA DA & 71.4                        & {62.3}          & 83.3          & 55.3          & 53.2     \\ 
 4 & D3ST + SGD-X DA (oracle) &
  73.8 & 69.7 & \textbf{92.5} & 62.1 & 27.9 \\ \hline 
 5 & D3ST-Turn (ours) &
 {\color[HTML]{000000} \textbf{75.9}} & 
69.5 & 
88.5 & 
63.2 & 
36.6  \\

 6 & D3ST-TurnSlot (ours) & 74.7 & \textbf{72.0} & 90.7 & \textbf{65.6} & \textbf{23.7} \\ \bottomrule 
\end{tabular}
}
\caption{Grounded D3ST models outperform strong baselines in both JGA and SS. Seen and unseen numbers decompose the SGD-X JGA (Section \ref{sec:dataset-metrics}),  \textit{oracle} indicates a model trained on SGD-X. "+" marks data augmentation (DA) during finetuning, and is followed by augmentation method name. Column maximum is in {\bf bold}. In all tables, numbers are averages of three runs.}
\label{tab:baselines}
\end{table}
Augmenting the finetuning dataset with the SGD-X prompts leads to a $13.2\%$ improvement in D3ST SGD-X JGA (\#1 vs \#4, Table \ref{tab:baselines}). In contrast, augmentation with synthetic prompts obtained through backtranslation or word-level augmentation improves performance by just $5.6\%$. On SGD, human-written prompts outperform the best performing synthetic ones (\#2) by a margin of $1.7\%$.

Gains obtained with synthetic prompts reflect some degree of lexical and syntactic diversity in the generated paraphrases. For example, a backtranslation of \textit{The amount of money to transfer} is \textit{Amount to be remitted} and \textit{The account type of the user} is backtranslated as \textit{User's account type}. The entailment scores (Table \ref{tab:schema-generations-metric}) show that backtranslation largely preserves the semantic content of the prompts. Meanwhile, if more edit operations are applied via EDA, the synthetic prompts are less faithful, as demonstrated by the sharp entailment decrease for the \textit{v4 \& v5} variants. Robustness did not improve when we experimented with a larger backtranslation-augmented dataset (Appendix \ref{appdx:larger-backtranslation-experiment}).
\begin{table}[tb]
\centering
\small
\begin{tabular}{ccllllc}
\toprule
  Schema &
  \multicolumn{1}{c}{v1} &
  \multicolumn{1}{c}{v2} &
  \multicolumn{1}{c}{v3} &
  \multicolumn{1}{c}{v4} &
  v5 \\ \hline

\multirow{3}{*}{} 
                             Backtranslation & 97.5                  & 96.5 & 95.9 & \multicolumn{1}{c}{-} & -                        \\
                             EDA             & 99.1                  & 98.5 & 96.6 & 93.2                  & \multicolumn{1}{l}{86.4} \\
                             SGD-X           & 89.7                  & 88.0 & 88.4 & 86.8                  & 87.5                     \\ \bottomrule
\end{tabular}
\caption{Semantic similarity of SGD and schema variants, measured by entailment \cite{narayan-etal-2022-well} }
\vspace{-0.2cm}
\label{tab:schema-generations-metric}
\end{table}

The SGD-X schemata "do not fully semantically overlap with the input as traditional paraphrasing requires" \cite{DBLP:conf/aaai/0001GR0ZW22}. This is consistent with SGD-X schema variants attaining lower entailment compared to backtranslated ones (Table \ref{tab:schema-generations-metric}). The annotators used the wider context of the service and common-sense knowledge to create diverse, high quality, schemas. Meanwhile, D3ST learns to identify slots using the uniform linguistic patterns of the SGD schema and it is not robust to the wide variety of styles annotators used in SGD-X. D3ST trained with augmented data via EDA or backtranslation improves compared to D3ST trained on SGD alone, but the large performance gap to human-written prompts indicates that strict paraphrasing introduces less diverse, task-relevant, cues in the prompt compared to humans.
\subsection{Prompt grounding with turns}
\label{sec:pgTurn}
Compared to D3ST, D3ST-Turn achieves absolute gains of $13\%$ and $6.1\%$ on SGD-X and SGD, respectively (\#1 vs \#5, Table \ref{tab:baselines}). D3ST + SGD-X DA outperforms D3ST-Turn on the seen services because it has been trained with these prompt paraphrases, whereas our model \textit{does not} see these prompts during training. Our model generalises more robustly as demonstrated by the $1.1\%$ (\#4 vs \#5) JGA improvement on unseen SGD-X services.

Our results show that grounding the model in knowledge-seeking turns, communicated before a slot is mentioned, addresses weaknesses of data augmentation (DA) with synthetic prompts. Such turns reflect how humans use the language in conversation when they communicate slot values, and may help the model more readily identify the relevant context for value extraction. This approach generalises well to unseen domains and is robust: we outperform all baselines on SGD and closely match the performance of augmentation with human-written paraphrases on SGD-X.

Descriptions and knowledge-seeking turns are complimentary: the latter can be thought of as an \textit{example} that could help the model interpret descriptions unseen at training time. Concretely, consider the \textit{messaging} domain, unseen in training. To extract the name of a location sharing recipient, a model evaluated on SGD-X ($v5$) is prompted with the description \textit{Name from address book}. Because T5 is pre-trained in a self-supervised way, without domain-specific finetuning, it may fail to identify that the aforementioned description refers to the name of a person: \textit{address book} never appears in the SGD training corpus. By attending over \textit{Who is the sharing recipient} and the description jointly, the model could interpret the description as referring to a person name. During training, the model has learned to identify names, for example, when predicting the value of the slot \textit{stylist name}. Indeed, the D3ST-Turn JGA in this domain is $41.8\%$ while the oracle model achieves just $28.5\%$. Our positive results may thus arise due to knowledge-seeking turns facilitating knowledge sharing between slots seen in training and unseen ones. 

\subsection{Prompt grounding with turns and slots}
\label{sec:pgTurnSlot}
D3ST-TurnSlot achives a $2.5\%$ gain on SGD-X compared to D3ST-Turn, outperforming the human-written prompts (\# 4 vs \#5, Table \ref{tab:baselines}). We posit that this is due to the high quality annotations SGD-X provides. This hypothesis is motivated by our empirical observation that slot names in SGD-X can contain more information compared to SGD ones. For example, the slot \textit{private visibility} in SGD is annotated as \textit{private visibility yes or no} in SGD-X ($v5$), which cues the model on which values should be generated for this slot. Also, in SGD-X, the slot names and descriptions may be complimentary. For example, the slot \textit{clock time of alarm} is described as \textit{Time for which the alarm is set} (SGD-X), whereas in SGD the equivalent slot name, \textit{alarm time}, is described as \textit{Time of the alarm}. On its own, the SGD description could refer to both an alarm to be created or an existing alarm, whereas the SGD-X description unambiguously identifies the slot as referring to an existing alarm.

D3ST-TurnSlot outperforms D3ST + SGD-X DA by $0.9\%$ on SGD (\#4 vs \#6, Table \ref{tab:baselines}) but lags behind D3ST-Turn by $1.1\%$. This confirms our earlier observation that, in SGD, unlike in SGD-X, slot names may not provide information about the slot that is not already contained in the description. We also find that there are slot name ambiguities across the SGD train and test sets. For example, the \textit{location} slot in the training set refers to cities, whereas in the test set it refers to addresses. This finding correlates with the study of \citet{DBLP:journals/corr/abs-2201-08904}, the D3ST authors, who find that lack of information in slot names and ambiguity lead to degraded JGA (on both SGD and SGD-X) of slot-name driven models compared to D3ST\footnote{A slot-driven model uses slot names instead of descriptions. For our example in Figure \ref{fig:D3STFigure} the equivalent prompt is \textit{0=name, 1=city, i1) get weather}. We refer the reader to Sections 4.3, 4.4 and 4.6 in \citet{DBLP:journals/corr/abs-2201-08904} for detailed comparisons of the effectiveness of these competing approaches.}. Our positive results on SGD-X show that combining the two sources of information can improve model robustness if they are complimentary and unambiguous.

\subsection{Grounded prompt ensembling}
\label{sec:pe}

\begin{table}[tb]
\centering
\Large
\resizebox{\columnwidth}{!}{
    \begin{tabular}{ccccccc}
\toprule
Model &
  SGD &
  SGD-X &
  Seen &
  Unseen &
  $\text{SS} \downarrow$ \\ \hline
D3ST-Turn& 
{\color[HTML]{000000} \textbf{77.2} \deemph{ 1.4}} & 
71.7 \deemph{ 2.2} & 
90.8 \deemph{ 2.3}& 
65.4 \deemph{ 2.2}& 
28.1 \deemph{ 8.5}\\
D3ST-TurnSlot &
  75.0 \deemph{0.3}&
  72.8 \deemph{0.8} &
   91.4 \deemph{0.7} &
66.6 \deemph{1.0} &
  \textbf{19.2} \deemph{4.5} \\ \bottomrule
\end{tabular}
}
    \caption{GPE improves SGD/SGD-X performance. Faded numbers are absolute improvements relative to the single pass models in Table \ref{tab:baselines} in rows \# 5 \& \# 6.}
    \label{tab:pe3-results-table}
\end{table}
We apply GPE by running three inference calls with distinct but semantically equivalent grounded prompts. We take the most common generation as the prediction. Table \ref{tab:pe3-results-table} shows significantly improved robustness compared to single-prompt decoding. Interestingly, D3ST-TurnSlot is improved  less compared to D3ST-Turn on both SGD and SGD-X owing to its significantly smaller prompt sensitivity ($23.7$ compared to $36.6$, Table \ref{tab:baselines}, \# 5 vs \#6). This shows that slot names increase the confidence of the model in its predictions, which may explain why we found D3ST-TurnSlot to slightly outperform D3ST-Turn (Section \ref{sec:pgTurnSlot}).

\subsection{Comparison with other models}
\label{sec:comparisons}
\begin{table}[tb]
\centering
\resizebox{\columnwidth}{!}{
\begin{tabular}{cccccc}
\toprule
Model &
  SGD  &
  SGD-X &
  Seen &
  Unseen &
  $\mbox{SS}\downarrow $ \\ \hline
 T5DST           &     70.0  & 50.4 & 58.5 & 47.7 & 87.0 \\ 
 MT-SGDST        &     80.1  & 60.8 & 72.5 & 56.9 & 69.5  \\          
 SDT-Seq         &     76.3  &  -   &   -  &  -   & -  \\ 
 SDT-Ind         &     {\textbf{78.2}}  &  -   &   -  &  -   & -  \\ \hline
D3ST-Turn (ours) &
 {\color[HTML]{000000} {75.8}} & 
69.5 & 
88.5 & 
63.2 & 
36.6  \\

D3ST-TurnSlot (ours) & 74.7 & \textbf{72.0} & 90.7 & \textbf{65.6} & \textbf{23.7} \\ \bottomrule 
\end{tabular}
}
\caption{SOTA DST models on SGD and SGD-X. Bottom rows repeated from Table \ref{tab:baselines} for easy comparisons.}
\label{tab:comparison-w-baselines}
\end{table}
We compare D3ST-Turn/TurnSlot with SOTA DST models (Table \ref{tab:comparison-w-baselines}). \textbf{T5DST} \cite{DBLP:conf/aaai/0001GR0ZW22} generates a slot value when prompted with a dialogue concatenated with a single description. The state is predicted \textit{iteratively} by prompting the model with each description. \textbf{MT-SGDST} \cite{DBLP:conf/interspeech/KapelonisGP22} uses semantic annotations, state history and handcrafted features with a multi-head BERT model for iterative prediction. 
\textbf{SDT-Seq} \citep{DBLP:conf/naacl/Gupta0Z0RW22} grounds the state tracker in a prompt containing a dialogue and its target state sequence and, like D3ST, predicts  the entire state in a single pass. \textbf{SDT-Ind} is an iterative version of SDT-seq, using annotated turns as prompts. 

Grounding prompts in knowledge-seeking turns makes D3ST competitive with SOTA approaches, significantly outperforming T5DST. MT-SGDST is better on the SGD but degrades significantly on SGD-X. Because it uses state histories and semantic annotations instead of system turns, this model suffers from large performance variability (Appendix \ref{sec:additional-details}): the difference between max and min SGD-X JGA across three runs is $11.8\%$ for this model but just $1.1\%$ for D3ST-Turn.

D3ST-Turn achieves $75.8\%$ on SGD, which is comparable with SDT-Seq ($76.3\%$). Our model is faster to train and decode owing to reduced prompt lengths and predicting a shorter state sequence\footnote{SDT predicts all slots, including inactive ones.}. SDT-Ind is better because it is prompted to return the value of each slot iteratively, with an example of how that typical slot occurs in conversation. GPE is cheaper and reduces the performance gap between SDT-Ind and D3ST-Turn to just $1.0\%$.

In terms of human effort, our approach is more scalable than, or comparable to, recent work SDT uses entire annotated dialogues or annotated turns as prompts (Figure 1, \citet{DBLP:conf/naacl/Gupta0Z0RW22}). These are defined by developers for unseen services, which is comparable to writing turns for each slot. Zero-shot transfer learning \citep{DBLP:conf/acl/CampagnaFML20} requires knowledge-seeking turns for generating synthetic dialogues\footnote{The \textit{direct questions} defined by \citet{DBLP:conf/acl/CampagnaFML20} are KSTs. See examples here: \href{https://bit.ly/3yvgGqi}{https://bit.ly/3yvgGqi}.} used to bootstrap DST models for new services. However, constraining entire dialogue generation is non-trivial and handcrafted grammars are required for each domain. This is very difficult to apply to the setting we consider, due to the large number of domains and complex multi-domain dialogue flows. We show that robust generalisation to new services can be achieved with few knowledge-seeking turns per slot which can be selected from the training corpus during finetuning and written by the developers for new services.

\begin{figure*}[]
    \centering
    \includegraphics[width=0.9\textwidth]{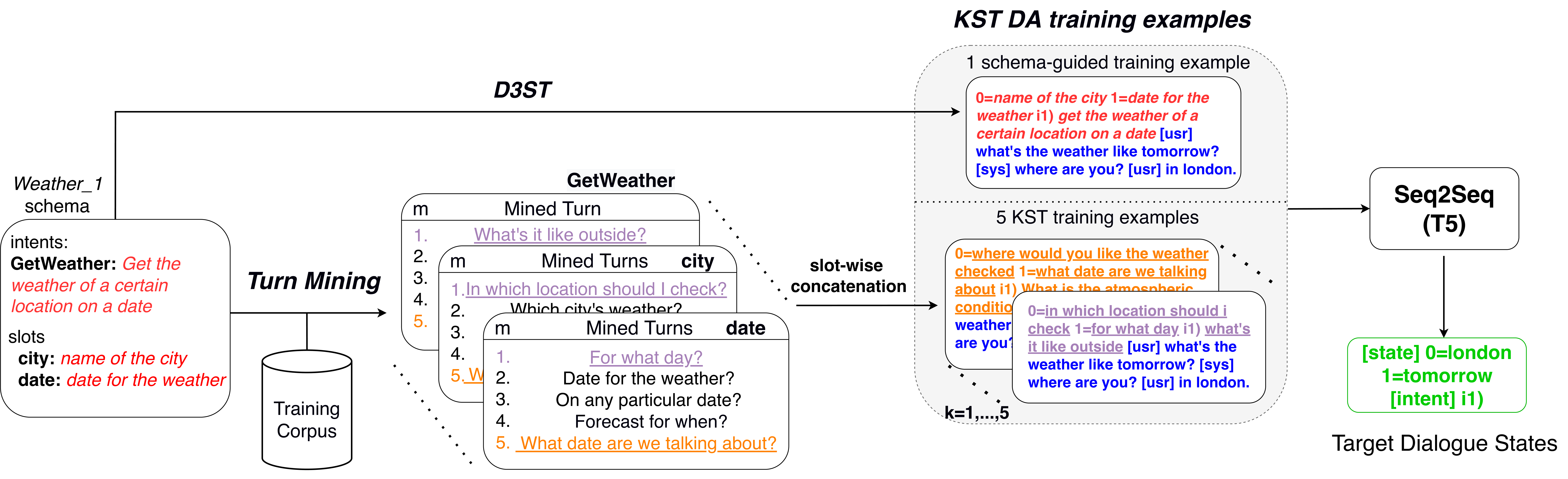}
    \caption{Data preprocessing pipeline for KST augmentation.  KST training examples share the conversation with the schema-guided training example, and concatenated KSTs, {\ul underlined}, replace the schema-guided prompt. }
    \label{fig:KST6x}
\end{figure*}
\subsection{Scaling behaviour}
\label{sec:scalability}
\begin{table}[tb]
\centering
\resizebox{\columnwidth}{!}{
\begin{tabular}{cccccc}
\toprule
Model &
  SGD  &
  SGD-X &
  Seen &
  Unseen &
  $\mbox{SS}\downarrow $ \\ \hline
  D3ST                 &    76.0           & 69.2              & 86.8              & 63.3          & 38.5 \\ 
 D3ST + SGD-X  DA             &    \textbf{77.4}  & {75.6}  & \textbf{93.2}     & 69.7          & \textbf{19.8} \\    \hline      
 D3ST-TurnSlot (ours)  &     \textbf{77.4} & \textbf{76.0}     & {92.6}  & \textbf{70.5} & { 20.5}\\ 
 \bottomrule 
\end{tabular}
}
\caption{Prompt grounding improves T5-large D3ST.}
\label{tab:t5-large}
\end{table}
Larger models have enhanced language understanding and common sense knowledge \cite{DBLP:journals/jmlr/RaffelSRLNMZLL20, DBLP:conf/iclr/ZhouLSL021}, reflected in the $12.7\%$ improvement of the baseline T5-large D3ST performance on SGD-X compared to its T5-base counterpart (\#1, Tables \ref{tab:baselines} and \ref{tab:t5-large}). Exposing the model to diverse prompts is still important, as demonstrated by the improved JGA of D3ST + SGD-X DA. We find that D3ST-TurnSlot matches the SGD performance and achieves a slight improvement on SGD-X ($0.6\%$), demonstrating that our approach scales to larger language models.

\subsection{Data augmentation or prompt grounding?}

In Section \ref{sec:pgTurn}, we discussed that knowledge-seeking turns may facilitate knowledge sharing between seen and unseen slots. We now investigate whether jointly encoding the turn and description by including them in the same prompt is the only way to impart this property or whether this can be achieved by augmenting the training data with prompts containing \textit{only} knowledge-seeking turns.

Figure \ref{fig:KST6x} shows our experimental setup for augmentation with knowledge-seeking turns. We sort the mined turn lists for each slot from Section \ref{sec:turn-mining} according to their Jaccard distance to the corresponding SGD schema description. We create $k=5$ \textit{increasingly diverse} prompts by replacing all the slot descriptions in a schema-guided training example with the $k$th knowledge-seeking turn. The resulting finetuning set is the same size as SGD-X.
\begin{table}[tb]
\centering
\resizebox{\columnwidth}{!}{
\begin{tabular}{ccccccc}
\toprule
\# & Decoding Prompt &
  SGD  &
  SGD-X &
  Seen &
  Unseen &
  $\mbox{SS}\downarrow $ \\ \hline
1 & Turn  \deemph{[GPE]} & 74.9 \deemph{[76.7]} & 71.7  \deemph{[73.9]} &  91.9 \deemph{[92.1]} &  65.0 \deemph{[67.7]}&  30.7 \deemph{[22.6]} \\          
2 & TurnSlot \deemph{[GPE]}   &  73.8 \deemph{[75.3]} & 71.0 \deemph{[72.9]} &  91.0 \deemph{[91.7]} &  64.3 \deemph{[66.6]} &  31.2 \deemph{[23.1]} \\ \hline

3 & D3ST    &  74.4 & 66.7 &  88.8 &  59.4 &  43.4  \\ \bottomrule 

\end{tabular}
}
\caption{JGA of D3ST with KST augmentation decoded with different prompt formats. GPE further improves these models (improvements inside brackets).}
\label{tab:knowledge-access-table}
\end{table}

Comparing the performance of augmented and grounded models (\#5 \& \#6, Table \ref{tab:baselines}  vs \#1 \& \#2, Table \ref{tab:knowledge-access-table}) shows that grounding D3ST is slightly more effective on SGD. 

On SGD-X, decoding the KST-augmented D3ST with TurnSlot prompt format causes a small ($1\%$) regression with respect to D3ST-TurnSlot, possibly due to mismatch between the train and test prompt formats. "Turn" decoding slightly improves over D3ST-Turn. Hence, both grounding and data augmentation with knowledge-seeking turns are effective for robust DST. Training with augmented data, converges slower and is resource intensive. Moreover, we find that grounding is more effective for larger language models (Table \ref{tab:t5-large-da-vs-kst}).

\begin{table}[tb]
\centering
\resizebox{\columnwidth}{!}{
\begin{tabular}{ccccccc}
\hline
\# & Model &
  SGD &
  SGD-X &
  Seen &
  Unseen &
  $\text{SS} \downarrow$ \\ \hline
1 & D3ST-TurnSlot &
  \textbf{77.4} &
  \textbf{76.0} &
  92.6 &
  \textbf{70.5} &
  { 20.5} \\ 
 \hline
2 & D3ST + KST DA/D3ST &
  {\color[HTML]{000000} { 76.3}} &
  72.8 &
  92.5 &
  66.3 &
  26.0 \\
3 & D3ST + KST DA/Turn &
  76.1 &
  74.6 &
  \textbf{93.4} &
  68.4 &
  26.0 \\ 
4 & D3ST + KST DA/TurnSlot &
  75.8 &
  73.6 &
92.2 &
  67.4 &
  28.2 \\ \bottomrule 
\end{tabular}
}
\caption{Grounding prompts (\#1) is more effective compared to KST-augmentation (\#2 - \#4) for robust DST with T5-large (770M parameters).}
\label{tab:t5-large-da-vs-kst}
\end{table}

\subsection{Why is grounding more effective?}

Decoding the KST-augmented model with the SGD/SGD-X schemata alone (i.e., without grounding) leads to a decrease in the unseen performance (\#1 \& \#2 vs \#3, Table \ref{tab:knowledge-access-table}). Grounding the prompt in KSTs at decoding time is crucial for improved robustness. As discussed in Section \ref{sec:pgTurn}, these turns facilitate knowledge sharing between seen and unseen slots. Without them, the model cannot access knowledge encoded in its weights, and robustly predict the dialogue state when the prompts are too dissimilar to the training schemata (Table \ref{tab:oracle-var-comparison}). 
\begin{table}[h]
\centering
\resizebox{\columnwidth}{!}{
\begin{tabular}{ccccccc}
\toprule
  SGD  &
  SGD-X (avg) &
  v1 &
  v2 &
  v3 &
  v4 &
  v5 \\ \hline
  {\color[HTML]{000000} 0.43} &
  {\color[HTML]{000000} 5.03} &
  {\color[HTML]{000000} 1.07} &
  {\color[HTML]{000000} 0.67} &
  {\color[HTML]{000000} 3.35} &
  {\color[HTML]{000000} 11.3} &
  {\color[HTML]{000000} {8.76}}  \\ \bottomrule
\end{tabular}
}
\caption{JGA difference between KST-augmented models decoded with Turn and D3ST prompt formats (\#1 \& \#3, Table \ref{tab:knowledge-access-table}). SGD-X JGA broken down by variant. $v5$ schema is the most dissimilar to the SGD test schema.}
\label{tab:oracle-var-comparison}
\end{table}

\section{Conclusion}
Grounding D3ST and data augmentation with knowledge-seeking turns are effective for robust schema-guided DST. Both improve D3ST robustness by a large margin compared to strong baselines and yield similar benefits as training on large, diverse collection of human-written prompts. Our proposed approach is competitive with or outperforms other SOTA DST models on SGD and SGD-X. We have also showed how prompt engineering can be applied to boost model robustness through grounded prompt ensembling, a novel technique that uses a single model for ensembling.

\section{Limitations}
One limitation of our approach is our decision to select the turns from the training data manually rather than automatically. Selecting $k$-diverse turns automatically is possible but requires efficient implementations given the size of the corpus and the quadratic complexity of the naive algorithm in the number of candidate turns. Implementing such algorithms requires far more expertise and time commitment compared to ensuring the selected turns are diverse manually. Such an approach is described by \citet{DBLP:conf/acl/LeeINZECC22}. 

While not an issue for SGD or other large scale corpora, the diversity of the training corpus may influence the performance of our approach as extracting lower diversity turns is expected to limit robustness improvements. However, knowledge-seeking turns existing in small corpora can be used to query large, possibly unlabeled, conversational databases to ensure prompt diversity. We left a detailed study of the impact of prompt diversity to DST robustness to future work.

Finally, for practically implementing our approach for unseen services, we require the developers to provide few examples of knowledge-seeking turns. Our currently in progress work explores generation of such turns automatically with very large language models.

\bibliography{anthology,custom}
\bibliographystyle{acl_natbib}

\appendix

\section{Turn mining details}
\label{appdx:kst-details}
For $31$ out of the $214$ slots there are no knowledge-seeking turns or there are less than $5$ distinct knowledge seeking turns\footnote{Only $25$ of these slots are unique as some slots repeat across services.} in the dialogue corpus. These includes \textit{result} slots which are communicate by the agent upon user query, such as the name or time of an existing alarm in the \textit{Alarms\_1} service. These slots do not appear in state annotations. Moreover, SGD dialogue flows are generated by semantic-level interaction between two machines modelled using push-down automata \cite{DBLP:conf/aaai/RastogiZSGK20}. As such, not all dialogue flows are covered. For example, in the \textit{Alarm\_1} service the user always states the name of a new alarm and the time they want to set it for so the system never asks for what time the new alarm should be set.

We circumvent these issues with two simple strategies, which are applied depending on whether a slot has knowledge-seeking turns in other services or not. The majority of the slots fall in the former case.

\textbf{Turn copy} The \textit{only} knowledge seeking turn for the \texttt{fare} slot in \texttt{Buses\_1} service is \textit{Thanks for that, how much did it cost?}. However, price is a generic concept which appears in other services (e.g. \texttt{Events\_1}) so instead of reducing prompt diversity by always using this turn, we copy knowledge-seeking turns from other services. In this instance, \textit{How much did it cost?}, \textit{Ticket fare for each passenger?}, \textit{Price per ticket?} and \textit{What price?} are copied. This strategy is applied to all slots that appear in other services.

\textbf{Span selection} For just $8$ slots, a relevant span appearing before or after the slot value is selected from turns annotated with actions \texttt{INFORM(s=v)} or \texttt{CONFIRM(s=v)} where $s$ is a slot for which no knowledge-seeking turns exist and $v$ is its value. For example, there are no turns where the system or user ask for the seating class of an airline ticket.  We select the span \textit{class flight ticket} instead of a full turn from the system turn \textit{Please confirm an Economy \underline{class flight ticket} to NY, tomorrow.}. The semantic annotation of this turn is \texttt{CONFIRM(destination=NY), CONFIRM(date=tomorrow), CONFIRM(seating\_class=Economy), CONFIRM(passengers=1).}

\section{Experimental details}
\label{appendix:state-tracking}
\subsection{D3ST implementation}
\label{appdx:d3st-details}
We process the data as described by \citet{DBLP:journals/corr/abs-2201-08904} with the following differences, which were indicated by the paper authors upon private communication: (1) the indices are separated by the \texttt{=} symbol in both the inputs and the targets, (2) for categorical slots which take the \textit{dontcare} special value, our output contains \textit{slot\_index: dontcare} substring and we do not include the special value in the prefix and (3) we lowercase inputs and targets.

The examples are truncated to the last $1,024$ tokens on the input side for the baseline and discarded altogether for Turn/TurnSlot prompt formats\footnote{This is just around $0.05\%$ of the data.}. We optimise with the Adafactor optimizer and effective batch size $32$, starting from the initial weights \texttt{google/t5-v1\_1-base} published by \texttt{huggingface} \cite{DBLP:journals/corr/abs-1910-03771}. We interpolate the learning rate linearly between $0$ and $10^{-4}$ over the first $1000$ steps and keep it constant thereafter. We select the model by evaluating the development set joint goal accuracy (JGA) every $5000$ gradient updates, stopping the training if said metric fails to improve after $3$ consecutive evaluations. 

All results in Section \ref{sec:results} are averages of three runs initialised with different random seeds. For all experiments, we used the same hyperparameters and stopping criteria as just described, with the exception of training the D3ST + SGD-X DA and D3ST + KST DA experiments for T5-large (rows 2-5 in Table \ref{tab:t5-additional}) where we allow all runs 1 epoch of augmented data (each SGD conversation is seen $6$ times) due to limited computational budget. 

\subsection{MT-SGD implementation}
\label{sec:additional-details}
The numbers presented are averages of three runs. The first SGD run (JGA of $83.2\%$) is based upon a metric file received from the authors. We could only reproduce $82.7\%$ of the quoted number, but we include the higher number in our average. We trained the model using the publicly available code\footnote{See it here: \url{bit.ly/3j8sPwj} } twice more obtaining to obtain $77\%$ and $80.2\%$. On SGD-X the JGA range is between $54.6\%$ and $66.4\%$ across three runs. We selected the \texttt{best} checkpoint as indicated in the repository's instructions.

\section{Backtranslation experiment}
\label{appdx:larger-backtranslation-experiment}
We experiment with larger backtranslation datasets to see if finetuning D3ST on a dataset the same size as the SGD-X dataset (Section \ref{sec:motivate-exp}) can improve results. We created two more variants by backtranslating the SGD schema via French and Russian, as done by \citet{DBLP:conf/eacl/HuangLQP21}. 

Augmenting with these additional examples negatively impacts model robustness (Table \ref{tab:backtranslation-extra}).  This may arise because increasing the number of training examples significantly (Table \ref{tab:backtr-bleu-self-bleu}) does not increase the prompt diversity by a large margin, and so the training distribution of the prompts is closer to the training data. Creating a diverse collection of paraphrases via backtranslation is thus challenging, as it requires access to translation systems to high-difficulty languages. This is necessary, since, as shown in Table \ref{tab:backtr-bleu-self-bleu} (BLEU, column $2$) translating to high-resourced languages such as French yields paraphrases that are lexically more similar to the input and are not as effective in improving the model robustness. Meanwhile, translation to difficult languages leads to semantic errors which may harm DST. For example, \textit{Station where the bus is leaving from} is backtranslated to \textit{Bus departure/arrival station} and \textit{Station where the bus is going to} is backtranslated as \textit{bus station} (via Japanese).
\begin{table}[tb]
\centering
\resizebox{\columnwidth}{!}{
\begin{tabular}{cccccc}
\toprule
Size &
  SGD  &
  SGD-X &
  Seen &
  Unseen &
  $\mbox{SS}$ \\ \hline

{\color[HTML]{000000} 4x} &
  {\color[HTML]{000000} \textbf{72.1}} &
  {\color[HTML]{000000} \textbf{62.2}} &
  {\color[HTML]{000000} \textbf{84.0}} &
  {\color[HTML]{000000} \textbf{54.9}} &
  {\color[HTML]{000000} \textbf{53.1}} \\
6x                    & 71.5                        & 61.0          & 82.5          & 53.8          & 54.4          \\
 \bottomrule
\end{tabular}
}
\caption{SGD and SGD-X JGA with backtranslation datasets of different size. We repeat line 2 from Table \ref{tab:baselines} in the top row, for easy comparison}
\label{tab:backtranslation-extra}
\end{table}

\begin{table}[tb]
\centering
\small
\resizebox{\columnwidth}{!}{\begin{tabular}{ccccccc}
\toprule
Metric                     & Method            & v1 & v2   & v3   & v4   & v5   \\ \hline
\multirow{2}{*}{BLEU} &
  Backtranslation 4x &
  \multicolumn{1}{l}{36.4} &
  \multicolumn{1}{l}{26.01} &
  \multicolumn{1}{l}{18.9} &
  - &
  - \\
 &
  Backtranslation 6x &
  \multicolumn{1}{l}{51.3} &
  \multicolumn{1}{l}{37.2} &
  \multicolumn{1}{l}{29.5} &
  \multicolumn{1}{l}{23.4} &
  \multicolumn{1}{l}{18.2} \\ \hline
\multirow{2}{*}{self-BLEU} & Backtranslation 4x & -  & 49.3 & 41.7 & -    & -    \\
                           & Backtranslation 6x & -  & 55.3 & 49.7 & 44.6 & 39.6 \\ \bottomrule
\end{tabular}
}
\caption{Lexical diversity metrics of backtranslated prompts. self-BLEU measures diversity of $n$ sentences}
\label{tab:backtr-bleu-self-bleu}
\end{table}
By grounding the model in turns collected from the corpus, not only do we create diverse inputs, but we guarantee that these correctly represent fine grained semantics and by-pass the issues encountered when constructing prompts via paraphrasing.

\section{SGD results}
\label{tab:seen-unseen-breakdown}
In the main body we report the SGD JGA accuracy as an upper bound for the D3ST model robust accuracy. To make our tables readable, we do not include SGD performance breakdown by seen/unseen services in Section \ref{sec:results}. We include it in Table \ref{tab:seen-unseen-breakdown} to facilitate future comparisons
\begin{table}[tb]
    \centering
    \footnotesize
\resizebox{\columnwidth}{!}{
\begin{tabular}{cccc}
\toprule

  Model &
  SGD &
  SGD-Seen &
  SGD-Unseen
\\ \hline

  D3ST &
  69.8&
  { 92.8} &
  62.2 \\ 

  D3ST + SGD-X DA &
  {\color[HTML]{000000} 73.8} &
  {\color[HTML]{330001} 92.7} &
  {\color[HTML]{000000} 67.5} \\ 
  
  \hline 

D3ST-Turn     & {\color[HTML]{000000} \textbf{75.8}} & \textbf{92.9}         &  \textbf{70.1}         \\

D3ST-TurnSlot &
  74.7 &
  {\color[HTML]{000000} { 92.8}} &
 68.7 \\ \hline \hline

D3ST + KST DA/Turn     & { 74.9}               & 92.6 & { 69.0}\\

D3ST + KST DA/TurnSlot & 73.8 & 92.5 & 67.6   \\ 
D3ST + KST DA/D3ST& 74.4 & { 92.8} & 68.3 \\ \bottomrule
\end{tabular}
}
    \caption{Breakdown on SGD JGA into seen and unseen services JGA for models reported Tables \ref{tab:baselines} and \ref{tab:knowledge-access-table}. }
    \label{tab:sgd-seen-unseen}
\end{table}

\end{document}